\documentclass[11pt]{article}
\usepackage[preprint]{acl}
\usepackage{times}
\usepackage{latexsym}
\usepackage[T1]{fontenc}
\usepackage[utf8]{inputenc}
\usepackage{microtype}
\usepackage{inconsolata}
\usepackage{graphicx}
\usepackage{booktabs}
\usepackage{multirow}
\usepackage{amsmath,amssymb}
\usepackage{xspace}
\usepackage{tikz}
\usetikzlibrary{arrows.meta,positioning,fit}
\usepackage{cleveref}
\crefname{section}{\S}{\S\S}
\Crefname{section}{\S}{\S\S}
\crefname{appendix}{App.}{Apps.}
\crefname{table}{Table}{Tables}
\crefname{figure}{Figure}{Figures}

\newcommand{\ci}[2]{\,[#1,\,#2]}  

\title{Confidence-Gated Extraction of Handwritten Bengali Land Records\\
with a Fine-Tuned Vision-Language Model}

\author{Md Tanveer Hossain Munim, Rezwan Ahmed, Md Anwarul Karim, \\
  \textbf{Kyungnam Lee, Jae Young Yoo} \\
  AI Geolab Ltd}

\begin{document}
\maketitle

\begin{abstract}
Land records in Bangladesh are handwritten registers. They contain personal data and may not leave government offices, so project operators digitise them by typing each page by hand. We describe a system, now deployed for government use, that shortens this work. A vision-language model reads each scanned page and drafts the structured record, and a person checks the draft instead of typing it. The model has 3 billion parameters, is fine-tuned on about 130,000 labelled pages, and runs locally on a single 16\,GB graphics card. We evaluate two tasks: reading register index pages and extracting full records. On villages not seen in training, the model's field score (the mean of per-field scores) is 95\% and 97\%, and a reviewer would have to change 2--4\% of the characters. The strongest commercial model we tested, Claude Opus~5.5 given worked examples, reaches 79\% and 81\% on the same pages and leaves 10--12\% of the characters to change. The model's token probabilities also show which fields are likely to be wrong. With a fixed threshold on them, 53\% of index pages go to a quick check, and 99.3\% of those pages are fully correct. Training on explanations written by a commercial model did not raise the score, and neither did a model with 7 billion parameters in the settings we could run. In a timing study, operators worked 3.0 (index) and 2.9 (full records) times faster with the system than by typing, without loss of accuracy.
\end{abstract}

\section{Introduction}
\label{sec:intro}

In Bangladesh, ownership of land is recorded in the Record of Rights, or \emph{khatian}. These are handwritten registers that contain personal data and may not leave government offices. Each khatian lists the owners of a holding, their shares and the numbers of the plots in it. The Department of Land Records and Surveys does not have the staff to digitise a century of these records, so the work falls to projects whose operators type each page by hand. Millions of pages have been scanned \citep{dailystar2017landdigital}, but a scan cannot be searched, cross-checked or updated until its contents are typed as structured data.

Automatic reading could shorten this work, but the pages are hard for general-purpose tools. Many clerks wrote the registers in Bengali over several decades, and several Bengali digits are easily confused. The output must also be exact, since one wrong digit in a plot number attaches land to the wrong record. KhatianDoc \citep{hasan2026khatiandoc}, a recent benchmark on the same document type, found that six multimodal large language models reach at most 25.8\% F1 on field extraction when used without task-specific training. A government office adds its own constraints. The hardware is one consumer graphics card, production pages may not leave the premises, and a person stays accountable for every record. To our knowledge, no earlier work has tested whether a small open vision-language model (VLM), fine-tuned on in-domain pages, is usable under these conditions.

This paper describes such a system. The model drafts each record, and a member of staff checks and corrects the draft before it is approved. Field accuracy alone cannot tell us whether this saves work. We therefore ask four further questions: how the system compares with commercial models, how much text a reviewer must still correct, whether the model's confidence identifies the drafts that need a careful check, and how much time operators save.

We fine-tune the open model Qwen2.5-VL-3B \citep{bai2025qwen25vl} with QLoRA \citep{dettmers2023qlora}, a memory-efficient fine-tuning method. We study two tasks from one district. One is reading the khatian and plot numbers from register index pages, and the other is extracting the full content of khatian records. On a 200-page sample per task from villages held out of training, the model's field score (the mean of per-field scores; \cref{sec:data}) is 95.0\% and 96.8\%. Claude Opus~5.5 \citep{anthropic2026opus55} with worked examples, the strongest of twelve baselines, reaches 78.7\% and 81.2\%. Our contributions are:
\begin{itemize}\setlength\itemsep{0pt}
\item A deployed system for drafting handwritten land records under government constraints, and the design rules that follow from those constraints (\cref{sec:system}).
\item An evaluation on unseen villages, with paired intervals over villages and a reviewer edit rate (\cref{sec:data,sec:results}).
\item Evidence that the lowest token probability within a field flags wrong fields well enough to decide how much review a page needs, while the first token's probability does not (\cref{sec:confidence}).
\item A timing study with operators (\cref{sec:timing}) and an analysis of design choices, including approaches that did not help (\cref{sec:lessons}).
\end{itemize}

\section{Related Work}
\label{sec:related}

\paragraph{Document extraction.} Donut \citep{kim2022donut} maps a document image directly to structured output, with no separate optical character recognition (OCR) step of the kind layout-aware models use \citep{huang2022layoutlmv3}. General VLMs now do the same when instructed \citep{bai2025qwen25vl}, and fine-tuned VLMs transcribe pages at scale \citep{poznanski2025olmocr}. CHURRO, a fine-tuned 3B Qwen2.5-VL, transcribes historical handwriting better than Gemini~2.5 Pro \citep{semnani2025churro}. Standard extraction benchmarks are mostly printed forms and receipts \citep{jaume2019funsd,huang2019sroie,park2019cord}. Handwritten forms have been studied for historical records \citep{wolf2025cm1,constum2025daniel} and French land registers \citep{tual2025landregistry}. Bengali resources cover handwritten digits and lines \citep{alam2018numtadb,rahman2022bnhtrd}, document layout \citep{shihab2023badlad}, OCR \citep{ali2023gold,islam2026bornodrishti} and government forms \citep{azad2026bafco}. The closest work is KhatianDoc \citep{hasan2026khatiandoc}, which evaluates untrained models on 107 khatians and reports low accuracy for all of them. Its data were not available to us. We instead train on about 130,000 pages and test on unseen villages.

\paragraph{Extraction with human review in production.} Deployed extraction systems commonly have a person check the output \citep{jagatap2025rxlens,publicfunding2025,pss2025,lofi2024,dua2025flexdoc}. Arctic-TILT decides which documents need review from the \emph{minimum} confidence over extracted values \citep{arcticTILT2025}. We adopt this rule and evaluate it for handwritten digit strings, village by village, with thresholds checked on held-out villages. Sending uncertain outputs to people is a form of selective prediction, in which a model abstains when its confidence is low \citep{elyaniv2010foundations,geifman2017selective}. For VLM extraction, ConfBench finds the probability of a field's first token a better confidence signal than the mean over its tokens \citep{roy2026confbench}. On our digit strings we find the opposite.

\section{Task, Data and Metrics}
\label{sec:data}

\paragraph{Tasks.} The two tasks correspond to two digitisation programmes. In task~A, the \emph{index} task, the system reads one index page of a Revisional Survey (RS) khatian register and returns the khatian number and the list of plot numbers. Index pages also carry five location fields, which are the same for every page of a register: the village or \emph{mouza}, its jurisdiction list (JL) number, the sub-district or upazila, the district and the division. In task~B, the \emph{full-content} task, the system reads one khatian record, always a single page. It returns the record's number, the owners' details as free Bengali text (names, fathers' names and addresses), the owners' share, and the plot numbers from three successive land surveys: the Cadastral Survey (CS), the State Acquisition survey (SA) and the RS. Task~B includes both original RS records and the later mutation records that document transfers. The khatian number, plot lists, share and owner details are the \emph{content fields}. Location fields come from a catalogue of registers (\cref{sec:lessons}), and the four area fields of task~B are reported separately (\cref{app:fields}).

\paragraph{Data.} All pages come from one district. The labels are existing manual transcriptions, typed by data-entry staff during earlier digitisation work. Each mouza (village) has its own registers. We assign whole mouzas to the training, validation or test set (80/10/10 by pages), so no test village appears in training. Task~A has 51,156 pages from 143 mouzas: 40,921 for training and 5,122 from 15 mouzas for testing. Task~B has 112,572 records from 59 mouzas: 89,357 for training and 10,509 from 9 mouzas for testing (9,277 RS and 1,232 mutation records). Pages without a scan are excluded.

\paragraph{Metrics.} We score each content field and average the scores over the content fields of a task, which gives the \emph{field score}. A single number, such as a khatian number, scores 1 if it is exactly correct and 0 otherwise. A list of plot numbers is scored by the F1 of the predicted set against the reference set. Owner details are scored with $1-\text{CER}$, where the character error rate (CER) is the edit distance to the reference divided by the reference length, capped at 1. We also report the average normalised Levenshtein similarity \citep[ANLS;][]{biten2019anls}. \emph{Page exact match} is the share of pages on which every field is exactly correct. Since the system produces drafts for review, we also measure the reviewer's \emph{edit rate}, adapted from translation edit rate \citep{snover2006hter}. This is the number of character edits needed to turn the draft into the reference, divided by the number of characters needed to type the reference from scratch, pooled over all fields and pages (\cref{app:metrics}). The edit rate is only a proxy for effort, so \cref{sec:timing} measures time.

\paragraph{Statistics.} Pages from one village share handwriting and register style, so all confidence intervals (CIs) are 95\% bootstrap intervals that resample whole villages \citep{field2007cluster}. Comparisons between systems are paired. Both are scored on the same pages, and the bootstrap is applied to the difference \citep{koehn2004paired}. The test sets contain only 15 and 9 villages, so we also report sign tests across villages. Our model is evaluated on the full test sets. Comparisons with baselines and ablations use a fixed sample of 200 test pages per task, stratified by village (\cref{app:metrics}). Most ablations are single training runs, so their CIs cover test-sample variation only. Repeated runs of one configuration differed by up to 0.8 points.

\section{System Design}
\label{sec:system}

\begin{figure}[t]
\centering
\begin{tikzpicture}[
  box/.style={draw, rounded corners=2pt, align=center, font=\scriptsize, inner sep=2pt, minimum height=7mm},
  arr/.style={-{Latex[length=1.5mm]}, thick},
  lab/.style={font=\scriptsize\itshape, inner sep=1pt}]
\node[box, text width=11mm] (scan) at (0,0) {Scanned page};
\node[box, text width=20mm] (vlm) at (2.45,0) {3B VLM on a local GPU: draft + field confidences};
\node[box, text width=17mm] (reg) at (5.35,0) {Location fields from register catalogue};
\node[box, text width=19mm] (gate) at (5.35,-1.5) {All fields $\geq 0.99$, no warnings?};
\node[box, text width=13mm] (quick) at (2.75,-1.05) {Quick check};
\node[box, text width=13mm] (full) at (2.75,-1.95) {Full review};
\node[box, text width=13mm] (appr) at (0.35,-1.5) {Second-person approval};
\draw[arr] (scan) -- (vlm);
\draw[arr] (vlm) -- (reg);
\draw[arr] (reg) -- (gate);
\draw[arr] (gate.west) -- ++(-0.3,0) |- node[lab, pos=0.75, above] {yes} (quick.east);
\draw[arr] (gate.west) -- ++(-0.3,0) |- node[lab, pos=0.75, below] {no} (full.east);
\draw[arr] (quick.west) -- ++(-0.35,0) |- (appr.east);
\draw[arr] (full.west) -- ++(-0.35,0) |- (appr.east);
\end{tikzpicture}
\caption{Review workflow. Pages with every field confident and no warnings get a quick check, the rest a full review, and a second person approves every record.}
\label{fig:system}
\end{figure}
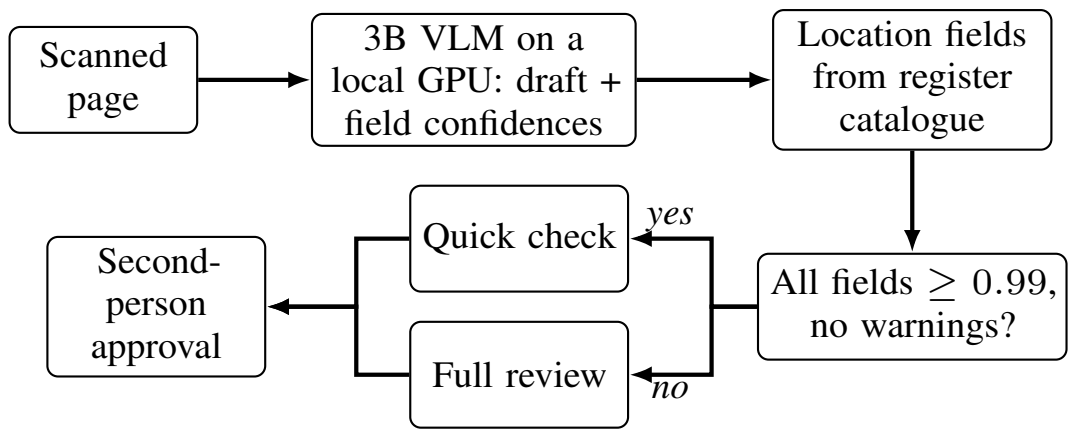

\paragraph{On-premises model.} The records contain personal data, so the model runs on a workstation inside the office. We use Qwen2.5-VL-3B, an open VLM that fits on one 16\,GB consumer graphics card. We fine-tune it with QLoRA \citep{dettmers2023qlora}, which stores the base model in 4-bit precision, keeps it fixed, and trains only small low-rank adapter matrices \citep{hu2022lora}. The model receives the page image and a short instruction and returns the fields as JSON. We train one model per task for one epoch, at an image budget of $1024\times28\times28$ pixels (about 0.8 megapixels). Training takes about 26 hours (A) and 42 hours (B) on the same card (\cref{app:training}).

\paragraph{Review and routing.} A member of staff reviews every record, with the scan next to the editable draft, and a second person approves it (\cref{fig:system}). For each field, the system records the lowest probability that the model assigned to any token of the field's value. We call this the \emph{field confidence}. A page goes to a \emph{quick check} when every field confidence is at least 0.99 and no validation rule, such as a check for duplicate plot numbers, has raised a warning. In a quick check, the reviewer sees the values in large type beside the scan and either confirms them with one key or sends the page to the full editor. All other pages get a full, field-by-field review. The validation rules can only add pages to full review, so \cref{sec:confidence} evaluates the confidence rule alone. We chose 0.99 when building the system, after a first look at held-out pages that included test villages. \Cref{sec:confidence} therefore also tests thresholds chosen on validation villages.

\section{Results}
\label{sec:results}

\paragraph{Baselines.} External services cannot be used in production, but with the data owner's permission we evaluated them to measure what they would offer (\cref{tab:main}). The commercial models are Claude Opus~5.5 \citep{anthropic2026opus55}, Gemini~3.1 Pro \citep{gemini31pro} and Gemini~2.5 Flash \citep{comanici2025gemini25}. The other models, run on a cloud service, are Llama~4 Maverick \citep{meta2025llama4}, Qwen3-VL-235B-A22B \citep{bai2025qwen3vl}, Mistral Large~3 \citep{mistral2025large3}, Gemma~3 27B \citep{gemmateam2025gemma3} and Nova Pro \citep{amazon2025nova}. On our own hardware we ran the untrained Qwen2.5-VL 3B and 7B, Gemma~4 26B-A4B \citep{gemmateam2026gemma4}, and a pipeline in which EasyOCR \citep{easyocr}, a general OCR system with Bengali support, reads the text and Qwen2.5-7B-Instruct extracts the fields. All systems receive the same instruction (\cref{app:prompts}), at first without examples (\emph{zero-shot}). Claude Opus~5.5, Llama~4 and Qwen3-VL also get a \emph{best-effort} setting, which adds two solved training pages as examples and sends the test page at higher resolution (2,400 instead of 1,600 pixels on the long side).
\begin{table*}[t]
\centering\footnotesize
\setlength{\tabcolsep}{2.6pt}
\begin{tabular}{@{}l rlr rlr rr@{}}
\toprule
& \multicolumn{3}{c}{A: index} & \multicolumn{3}{c}{B: full content} & \multicolumn{2}{c}{A / B} \\
\cmidrule(lr){2-4}\cmidrule(lr){5-7}\cmidrule(l){8-9}
System (setting) & Score & Gap [95\% CI] & Edit & Score & Gap [95\% CI] & Edit & \$/1k pages & s/page \\
\midrule
\multicolumn{9}{@{}l}{\emph{Run on our own hardware}} \\
\textbf{Ours} (3B, fine-tuned) & \textbf{95.0} & -- & \textbf{2.3} & \textbf{96.8} & -- & \textbf{4.1} & local$^\ast$ & 2.2 / 5.0 \\
Qwen2.5-VL 3B (zero-shot) & 3.3 & -- & -- & 23.0 & -- & -- & -- & -- \\
Qwen2.5-VL 7B (zero-shot) & 11.3 & -- & -- & 25.8 & -- & -- & -- & -- \\
Gemma 4 26B-A4B (zero-shot) & 17.1 & 77.9\ci{74.4}{81.4} & -- & 37.8 & 59.0\ci{55.3}{63.6} & -- & -- & 4.3 / 6.9 \\
EasyOCR$\rightarrow$Qwen2.5-7B & 9.0 & 86.0\ci{82.5}{89.5} & 91.1 & 29.4 & 67.3\ci{64.6}{70.2} & 65.6 & -- & 2.9 / 6.5 \\
\midrule
\multicolumn{9}{@{}l}{\emph{External services}} \\
Claude Opus 5.5 (zero-shot) & 75.0 & 20.0\ci{13.2}{26.5} & 10.6 & 75.4 & 21.4\ci{16.8}{27.1} & 13.2 & 18.6 / 26.3 & 5.3 / 8.9 \\
Claude Opus 5.5 (best effort) & 78.7 & 16.3\ci{9.1}{23.4} & 10.3 & 81.2 & 15.6\ci{11.5}{21.1} & 12.0 & 25.8 / 35.7 & 3.2 / 8.3 \\
Gemini 3.1 Pro (zero-shot)$^\dagger$ & 72.6 & 22.9\ci{16.6}{28.6} & 14.7 & -- & -- & -- & n/a & 12.5 / -- \\
Gemini 2.5 Flash (zero-shot) & 17.5 & 77.5\ci{71.5}{82.4} & -- & 50.2 & 46.6\ci{43.6}{50.3} & -- & -- & -- \\
Llama 4 Maverick (zero-shot) & 17.6 & 77.4\ci{71.7}{82.9} & -- & 37.9 & 58.8\ci{54.1}{65.2} & -- & 1.3 / 1.4 & -- \\
Llama 4 Maverick (best effort) & 36.3 & 58.7\ci{50.7}{65.7} & 52.7 & 48.9 & 47.8\ci{42.8}{54.9} & 41.4 & 3.6 / 3.8 & 8.6 / 7.6 \\
Qwen3-VL-235B (zero-shot) & 5.0 & -- & -- & 38.1 & -- & -- & 1.6 / 2.4 & -- \\
Qwen3-VL-235B (best effort) & 32.3 & 62.7\ci{55.9}{68.9} & -- & 47.2 & 49.6\ci{46.1}{54.0} & -- & 4.8 / 5.1 & 5.6 / 7.6 \\
Three others (zero-shot)$^\ddagger$ & 1.7--4.0 & $\geq$\,91 & -- & 19.8--27.1 & $\geq$\,69 & -- & 1.4--4.5 & -- \\
\bottomrule
\end{tabular}
\caption{Main results on the 200-page test samples (\%). \emph{Score}: field score. \emph{Gap}: how many points our model leads, with a 95\% CI; rows without a CI trail by 58 points or more. \emph{Edit}: share of characters a reviewer must change. \emph{\$/1k pages}: list price. \emph{s/page}: median time per request; for our model, time per page on one GPU. $^\ast$Electricity only, about 0.1 (A) and 0.25 (B) kWh per 1,000 pages. $^\dagger$Ran out of credits after 166 pages; its gap uses those pages. $^\ddagger$Mistral Large~3, Gemma~3 27B and Nova Pro. On the full test sets our model scores 95.2\ci{92.3}{98.2} (A) and 96.5\ci{95.9}{97.8} (B).}
\label{tab:main}
\end{table*}

\paragraph{Main comparison.} The fine-tuned 3B model scores higher than every baseline on both tasks (\cref{tab:main}), and every reported paired CI excludes zero. Our model is not behind in any test village. On task~A, it leads zero-shot Claude Opus~5.5, Gemini~3.1 Pro and best-effort Llama~4 in all 15 (sign test $p<10^{-4}$; smallest gap 3.8 points). It leads best-effort Claude Opus~5.5 in 14, with one tie ($p<0.001$). On task~B, it leads Claude Opus~5.5 and best-effort Llama~4 in all 9 ($p=0.004$; smallest gap 9.9). The services are cheap to use, so the case for a local model rests on accuracy and privacy.

\paragraph{Reviewer effort.} Edit rates order the systems the same way. A reviewer of our drafts changes 2.3\% (A) and 4.1\% (B) of the characters. For Claude Opus~5.5 at its best the figures are 10.3\% and 12.0\%, which is 3--4.5 times more, with paired differences of 8.0\ci{3.4}{12.9} and 8.0\ci{6.4}{9.8} points. The OCR pipeline leaves 91\% and 66\%. We did not test confidence signals for the services.

\paragraph{Errors by field.} On the full test sets, the khatian number is correct on 95.0\% (A) and 91.6\% (B) of pages. Owner details in task~B have a CER of 4.1\% (ANLS 96.0\%), and \cref{app:fields} gives every field. Most errors of all systems are misread digits, often between visually similar Bengali digits. On the task~A sample, one digit is misread in 11 of our 12 wrong khatian numbers and in 37 of best-effort Claude Opus~5.5's 49.

\section{Confidence-Gated Review}
\label{sec:confidence}

Every draft is reviewed, so the question is how much review each page needs. Confidence can answer this only if it ranks errors well. We measure that with the area under the receiver operating characteristic curve (AUROC): the probability that a randomly chosen correct field receives a higher confidence than a randomly chosen wrong one. A value of 0.5 is chance and 1 is perfect.

\begin{table}[t]
\centering\footnotesize
\setlength{\tabcolsep}{3pt}
\begin{tabular}{@{}lrrrrr@{}}
\toprule
& & \multicolumn{2}{c}{AUROC} & \multicolumn{2}{c}{Gate at 0.99} \\
\cmidrule(lr){3-4}\cmidrule(l){5-6}
Unit & Acc. & lowest & first & Passed & Acc. \\
\midrule
A: khatian number & 95.0 & 0.961 & 0.599 & 76.2 & 99.6 \\
A: plot list & 90.9 & 0.956 & 0.641 & 66.0 & 99.6 \\
A: whole page & 86.8 & 0.946 & 0.525 & \textbf{52.8} & \textbf{99.3} \\
\midrule
B: khatian number & 91.7 & 0.969 & 0.885 & 75.4 & 99.7 \\
B: RS plot list & 97.2 & 0.919 & 0.616 & 85.5 & 99.6 \\
B: share & 97.6 & 0.921 & 0.894 & 88.1 & 99.6 \\
B: owner details & 31.3 & 0.946 & 0.753 & 0.6 & 98.4 \\
B: page, numeric fields & 84.4 & 0.929 & -- & 51.1 & 98.9 \\
\bottomrule
\end{tabular}
\caption{How well confidence flags errors (full test sets, \%). \emph{Acc.}: share exactly right (plot lists: the whole list). \emph{AUROC}: from the lowest token probability in the field or from the first token's. A page's confidence is that of its least confident field. \emph{Numeric fields}: all except owner details. \emph{Gate at 0.99}: share of pages with every field at 0.99 or above, and their accuracy. The 11 pages without parsable output are left out, so accuracies can differ from \cref{tab:fields} by 0.1.}
\label{tab:conf}
\end{table}
\begin{figure}[t]
\centering
\includegraphics[width=\columnwidth]{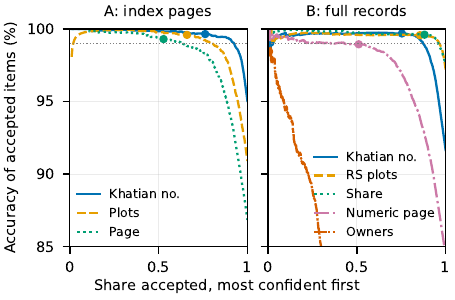}
\caption{Accuracy of the accepted fields or pages as more are accepted, most confident first (full test sets). Dots mark the 0.99 threshold, and the dotted line is 99\% accuracy. The owner-details curve continues below 85\%.}
\label{fig:coverage}
\end{figure}

\paragraph{Aggregating token probabilities.} The lowest token probability within a field separates correct from wrong fields well, with AUROCs of 0.92--0.97 (\cref{tab:conf}). The first token's probability works much worse. It is close to chance on task~A (0.53--0.64) and below the minimum for every field of task~B except the CS list, which is wrong on only 2 pages. A misread digit is often not the first one, so the first token says little about it. The product of token probabilities does as well as the minimum (page AUROC 0.949 vs 0.946), and the mean is slightly worse (0.927).

\paragraph{Fixed threshold.} We first evaluate the confidence part of the deployed rule, all fields at least 0.99. On the index task it sends 52.8\%\ci{42.7}{69.5} of pages to the quick check, and 99.3\%\ci{98.7}{99.8} of those pages are exactly correct on every field (\cref{fig:coverage}). Only 19 of the 2,702 pages it passes are wrong, and 97.2\% of the wrong pages go to full review. Accuracy varies between villages. In 5 of the 15, fewer than 99\% of the passed pages are correct (\cref{app:gate}).

\paragraph{Owner details.} On task~B, owner details almost never reach 0.99 (0.6\% of pages), so nearly every record gets a full review. In that review, staff check owner names against national identity records, and the draft saves them typing the names. For the numeric fields alone, the gate passes 51.1\% of pages at 98.9\%\ci{98.6}{99.4} accuracy.

\paragraph{Thresholds chosen on validation villages.} Because 0.99 was chosen with some test villages in view, we also chose thresholds on the 14 validation villages of task~A only and applied them to the test villages. The lowest threshold reaching 99\% on validation (0.965) passes 70.2\%\ci{58.7}{84.7} of test pages, but at only 98.7\%\ci{97.9}{99.4} accuracy. A threshold set by a statistical bound of 1\% error on validation (\cref{app:gate}) is 0.989, almost the deployed 0.99, and passes 54.5\%\ci{44.3}{71.2} at 99.2\%\ci{98.6}{99.7}. No rule we tested reaches 99\% in every village while passing a useful share of pages (\cref{app:gate}), so a new village should start with a stricter threshold and an audit of its quick-checked pages.

\section{Operator Timing Study}
\label{sec:timing}

Edit rates and gate coverage only suggest how much work is saved, so we also timed the work directly. Four operators each processed 30 test pages per task, drawn at random across the test villages. Each typed half of the pages from the scan, as in current practice, and checked drafts on the other half, routed by the confidence rule. On task~B, owner details were always fully reviewed. The halves were balanced on village and gate decision, and each page was typed by two operators and checked by the other two (\cref{app:timing}).

\begin{table}[t]
\centering\footnotesize
\setlength{\tabcolsep}{2pt}
\begin{tabular}{@{}llr@{}}
\toprule
Task & Pages & Speed-up [CI] \\
\midrule
A & all (30) & 3.0\ci{2.6}{3.5} \\
& quick (16) & 3.3\ci{2.8}{3.9} \\
& full (14) & 2.9\ci{2.4}{3.4} \\
\midrule
B & all (30) & 2.9\ci{2.5}{3.3} \\
& quick (15) & 4.7\ci{3.9}{5.7} \\
& full (15) & 1.9\ci{1.6}{2.4} \\
\bottomrule
\end{tabular}
\caption{How many times faster operators were when checking drafts than when typing, for all pages and by the route the gate chose. The 95\% CI resamples operators and pages.}
\label{tab:timing}
\end{table}

Operators were 3.0 times faster with the system on task~A and 2.9 times on task~B (\cref{tab:timing}). The gain is largest on pages sent to a quick check, 3.3 and 4.7 times. Pages sent to full review were also done faster, by 2.9 times on task~A and 1.9 on task~B. Accuracy did not fall. Against the reference transcription, every content field was right on 56 of 60 hand-typed pages and 59 of 60 checked drafts on task~A. On task~B, the counts were 14 and 37 of 60. They are low both ways because a page counts only if its long owner text matches exactly.

\section{Analysis and Ablations}
\label{sec:lessons}

\Cref{tab:ablation} compares design choices.

\begin{table}[t]
\centering\footnotesize
\setlength{\tabcolsep}{3pt}
\begin{tabular}{@{}lrr@{}}
\toprule
Variant (3B unless stated) & Score & $\Delta$ [95\% CI] \\
\midrule
\multicolumn{3}{@{}l}{\emph{Training data (A; B in brackets)}} \\
6k pages & 93.0 [92.5] & \\
20k pages & 94.7 [94.8] & $+$1.7\ci{$-$0.1}{4.4} \\
all (41k [89k] pages) & 95.0 [96.8] & $+$0.3\ci{$-$1.6}{3.0} \\
\midrule
\multicolumn{3}{@{}l}{\emph{Rationale distillation (6k pages)}} \\
A: + Gemini rationales & 92.6 & $-$0.4\ci{$-$1.9}{1.2} \\
B: + Gemini rationales$^\ast$ & 91.3 & $-$0.2\ci{$-$0.7}{0.3} \\
\midrule
\multicolumn{3}{@{}l}{\emph{Model size (A)}} \\
7B, 6k pages, budget 768 & 90.7 & $-$2.3\ci{$-$4.7}{0.0} \\
7B, 20k pages, 40\,GB GPU & 94.6 & $-$0.1\ci{$-$1.9}{2.1} \\
\midrule
\multicolumn{3}{@{}l}{\emph{Image budget, train $\rightarrow$ read (A, 6k pages)}} \\
1024 $\rightarrow$ 1024 & 89.9 & \\
1024 $\rightarrow$ 1280 (used) & 93.0 & $+$3.1\ci{0.5}{5.9} \\
1280 $\rightarrow$ 1280 & 89.3 & $-$0.6\ci{$-$3.0}{2.2} \\
\bottomrule
\end{tabular}
\caption{Effect of design choices (field score, \%, 200-page samples; one training run each unless marked). $\Delta$ compares a row with the row above (data), the same model trained without rationales (distillation), the 3B model trained on the same data (size), or the 1024$\rightarrow$1024 row (image budget, in units of $28\times28$ pixels). $^\ast$Mean of three training runs (\cref{app:training}).}
\label{tab:ablation}
\end{table}

\paragraph{Rationale distillation.} For 2,000 of the 6,000 training pages per task, we gave Gemini~2.5 Flash the page and its reference labels and had it write a short explanation, or \emph{rationale}, of where each field sits on the page and how it reads. We then trained the model to produce the rationale before the JSON \citep{hsieh2023distilling}. This did not help. It scored 0.4 (A) and 0.2 (B) points below plain training, and both CIs include zero (\cref{tab:ablation}). Generating the rationale also made inference 2--5 times slower. This agrees with findings that rationale training does not transfer reasoning \citep{wadhwa2024mysteries} and that step-by-step reasoning tends to hurt perception tasks \citep{jin2026looklight}.

\paragraph{Model size and resolution.} A 7B model did not score higher than the 3B model in either comparison we could run. Neither isolates model size: on the 16\,GB card the 7B model needed a smaller image budget, and on a 40\,GB cloud GPU it was trained without quantisation. Reading pages at a higher image budget than in training helped most. The same trained model scores 3.1\ci{0.5}{5.9} points higher, and it also leads on the validation pages (\cref{app:training}). Image classifiers show the same train--test resolution effect \citep{touvron2019fixres}.

\paragraph{Register metadata.} The five location fields are the same throughout a register, but they are written inconsistently or not at all on many index pages. Reading them limits page exact match to 7.6\% on task~A. The office knows which register it is scanning, so the system takes them from a catalogue of registers instead, which raises page exact match to 86.8\%\ci{79.2}{94.8}.

\section{Limitations}
\label{sec:limitations}

All data come from one district, so transfer to other districts and handwriting styles is untested. The same clerks may have written registers for several villages, so a held-out village does not guarantee unseen handwriting. The labels are earlier manual transcriptions whose error rate we did not measure, and part of the remaining error may be label noise. Baselines and ablations use 200-page samples, and most ablations are single runs (\cref{sec:data}). With 15 and 9 test villages, intervals that resample villages may be somewhat too narrow, but the sign tests do not depend on resampling. The OCR baseline uses a general recogniser, and one trained on handwritten Bengali would likely do better. The timing study is small and scored operators against a reference that is itself a transcription. We did not measure whether quick-check reviewers come to accept drafts without looking (automation bias) over long use. The second approval and the audits suggested in \cref{sec:confidence} are meant to limit it. Usage statistics from the deployment cannot be reported, at the data owner's request. The data cannot be released, which limits independent replication. We will release the evaluation code, prompts and split procedure upon acceptance.

\section{Deployment and Conclusion}
\label{sec:conclusion}

The system is deployed for government use, on one workstation with a 16\,GB graphics card (175\,W measured). It drafts about 1,650 index pages or 720 full-content records per hour, and it logs each draft and the reviewer's final version.

A 3B VLM fine-tuned on in-domain pages can draft handwritten Bengali land records on one consumer graphics card. It leaves 3--4.5 times fewer characters to correct than the best commercial model we tested and sends half of the index pages to a quick check. In our timing study, operators worked 2.9--3.0 times faster with it, without loss of accuracy.

\section*{Ethical Considerations}

The records contain personal data, and the data owner, a government body, permitted their use for this work. The system runs on premises. People review and approve every record, since an error can affect land rights. The system changes the operators' work from typing to checking, and a person remains accountable for each record. Pages left the premises only with the data owner's permission. For the baselines and the distillation experiment, they went only to paid services whose terms exclude training on customer data. These were Google's Gemini API (4,000 training pages for the rationales, 400 test pages for Gemini~2.5 Flash, and up to 200 test pages for Gemini~3.1 Pro), Amazon Bedrock (the same 400 test pages for the other models), and the Claude API (the same 400 test pages and the example pages, in three runs). The Claude API deletes inputs and outputs within 30 days and does not use them for training. On Bedrock, the account's data-retention setting was set to \emph{none} during these runs. One 7B training run used a rented cloud GPU, and its data were deleted along with the instance. The timing study ran on a closed local network. Operators gave informed consent, and results are reported without identifying them. The data and model cannot be released publicly. The model and the 200-page evaluation samples per task can be made available to authorised parties on a secure machine on request.

\bibliography{refs}

\appendix

\section{Prompts}
\label{app:prompts}

All systems receive the same instruction per task. Task~A:
\begin{quote}\small\ttfamily
You are an expert reader of handwritten Bangladeshi RS khatian (land-record) index pages. Read the single scanned page in the image and extract its index fields.
Return ONLY JSON with these keys: "khatian\_number": the khatian/account number (string of digits). "plot\_numbers": list of dag/plot numbers on the page (each a string of digits), in the order written. "jl\_number": the JL number if present, else "". "mouza", "upazila", "district", "division": the name in Bengali if present, else "".
Rules: Output digits as Western numerals (0-9), NOT Bengali numerals. Do not invent values. If a field is absent, use "" (or [] for plot\_numbers). Output JSON only, no explanation.
\end{quote}
The task~B instruction lists the keys khatian\_number, owners\_information (``the full owners text exactly as written in Bengali, including names and father/address parts, joined as one string''), percent\_share, rs\_plot, sa\_plot, cs\_plot and four area fields, with the same rules plus ``Keep Bengali script for the owners\_information text'' and ``A decimal like `.09' means less than one; keep the leading dot.'' The distilled models receive the same instruction followed by ``First, think briefly inside <think>...</think>: locate each field on the page and read its handwriting. Then output the JSON object after the </think> tag.''

\section{Training and Inference Details}
\label{app:training}

\paragraph{Fine-tuning.} Qwen2.5-VL-3B-Instruct, base weights in 4-bit NormalFloat (NF4) with double quantisation and bf16 computation. LoRA rank 16, $\alpha=32$, dropout 0.05, on the query, key, value, output, gate, up and down projections of the language model. Loss on the JSON answer only (instruction and image tokens masked). One epoch; learning rate $10^{-4}$ with cosine decay and 3\% warm-up; batch size 1 with 8 accumulation steps; 8-bit AdamW; gradient checkpointing; seed 42 (also 1 and 2 for the task~B distillation comparison, where the three runs score 91.1--91.9 without rationales, mean 91.5, and 91.1--91.7 with them). Image budget $1024\times28\times28$ pixels in training. The 7B comparison at full budget used bf16 LoRA (no quantisation) on a 40\,GB GPU.

\paragraph{Inference.} Greedy decoding, at most 1,280 new tokens, repetition penalty 1.1, image budget $1280\times28\times28$ pixels, base in 4-bit. The field scores were mostly obtained one page at a time. The confidence analysis, the per-field results (\cref{tab:fields}), the validation pages and the starred rows of \cref{tab:ablation} used batches of 8. Batching changes the full-test field scores by at most 0.1 points, but on the 200-page sample it moved the task~B 6k model from 92.5 to 91.9. On the 5,113 validation pages of task~A, reading at 1280 after training at 1024 again leads both 1024$\rightarrow$1024 ($+$3.2\ci{1.4}{4.5}) and 1280$\rightarrow$1280 ($+$1.3\ci{0.0}{2.3}). On the 16\,GB card, peak GPU memory in use was 3.2\,GiB for reading pages one at a time (2.5\,GiB for the 4-bit base with both task adapters) and 13.1\,GiB for training (task~B runs).

\paragraph{Baselines.} External services were called with the same instruction, images resized to 1,600 pixels on the long side (2,400 in the best-effort setting), and temperature 0 where the service accepts it. Claude Opus~5.5 does not, so it was run with its default sampling. Best-effort prompts contain two solved training pages as examples. Gemma~4 26B-A4B is a mixture-of-experts model with 3.8B of its 25.2B parameters active. A response without parsable JSON counts as an empty prediction. This affects at most 6.5\% of pages for every VLM baseline except Nova Pro (18--23\%) and the untrained Qwen2.5-VL models on task~B (14--19\%). The other systems therefore score low because they misread the page, not because their output could not be parsed. Gemini~3.1 Pro's 166 pages cover all 15 test villages of task~A. In the OCR pipeline, the reference khatian number appears in EasyOCR's output on at most 49\% (A) and 25\% (B) of pages. Even then, the text model picks it from about 30 digit strings per page only 32\% (A) and 14\% (B) of the time, since it does not see the layout.

\section{Per-Field Results}
\label{app:fields}

\Cref{tab:fields} gives the score of each field on the full test sets, including the four area fields of task~B, which are not part of the field score. Per village, the field score ranges from 81.7\% to 99.4\% on task~A (median 98.6\%; the lowest is an 18-page village) and from 92.1\% to 98.7\% on task~B. An owner text has about 150 characters and is rarely reproduced without any difference, so exact match understates its usefulness as a draft.

\begin{table}[h]
\centering\small
\begin{tabular}{@{}llr@{}}
\toprule
Task & Field & Score (\%) \\
\midrule
A & khatian number (exact) & 95.0 \\
A & plot list (set F1) & 95.3 \\
A & plot list (exact) & 90.9 \\
\midrule
B & khatian number (exact) & 91.6 \\
B & owner details ($1-$CER) & 95.9 \\
B & owner details (ANLS) & 96.0 \\
B & owner details (exact) & 31.3 \\
B & share (exact) & 97.5 \\
B & RS plot list (set F1) & 97.5 \\
B & SA plot list (set F1) & 96.8 \\
B & CS plot list (set F1) & 100.0 \\
\midrule
B & total area, acres (exact) & 99.8 \\
B & total area, decimals (exact) & 98.9 \\
B & owned area, acres (exact) & 99.9 \\
B & owned area, decimals (exact) & 87.1 \\
\midrule
B & field score, RS records (9,277) & 96.4 \\
B & field score, mutation records (1,232) & 97.4 \\
\bottomrule
\end{tabular}
\caption{Score of each field for our model on the full test sets. Every page counts, and a page without parsable output scores as an empty answer. Area fields are not part of the field score.}
\label{tab:fields}
\end{table}

\section{Confidence Gate by Village}
\label{app:gate}

On task~A, the fixed 0.99 gate passes between 16.7\% and 87.6\% of the pages of a village (median 63.7\%). The share of passed pages that are fully correct is 100\% in 7 villages, 99.2--99.8\% in 3, and 96.0--98.9\% in 5. Three of these five have fewer than 60 passed pages. The gate sends 97.2\% of all wrong pages to full review (94.5\% for the khatian number and 97.0\% for the plot list as single fields). For the numeric fields of task~B, the gate passes 0--71.5\% of the pages of a village. Passed pages are fully correct in 95.4--100\% of cases, and 4 of the 8 villages with passed pages fall below 99\%.

\Cref{tab:thresholds} shows the task~A page gate at other thresholds.

\begin{table}[h]
\centering\footnotesize
\setlength{\tabcolsep}{2.5pt}
\begin{tabular}{@{}lrrrr@{}}
\toprule
Threshold & Passed & Acc. & Villages $<$99\% & Wrong sent to full \\
\midrule
0.95 & 73.0 & 98.6 & 7/15 & 92.0 \\
0.98 & 62.7 & 99.0 & 7/15 & 95.4 \\
\textbf{0.99} & \textbf{52.8} & \textbf{99.3} & \textbf{5/15} & \textbf{97.2} \\
0.995 & 40.5 & 99.7 & 3/15 & 99.0 \\
0.999 & 12.6 & 99.9 & 1/15 & 99.9 \\
\bottomrule
\end{tabular}
\caption{The page gate on the task~A test set at other thresholds (\%). The system uses 0.99.}
\label{tab:thresholds}
\end{table}

The threshold with an error bound in \cref{sec:confidence} follows Learn-then-Test \citep{angelopoulos2021ltt}. It is the lowest threshold on a grid of 120 values whose one-sided 95\% Clopper--Pearson upper bound \citep{clopper1934} on the error of passed validation pages is at most 1\%, with a Bonferroni correction over the grid. On the validation villages, the 0.99 gate passes 46.6\% of pages at 99.7\% accuracy. With either threshold chosen on validation, 5 (bounded) or 7 (lowest reaching 99\%) of the 15 test villages fall below 99\%. The bound treats validation pages as independent, and accuracy varies between villages, so it does not carry over to each village. A bound with villages as the unit would be far looser with only 14 validation villages.

Choosing thresholds within the test set instead, for each village on the other 14, gives a similar picture. The lowest threshold reaching 99\% passes 63.9\%\ci{56.4}{77.3} of pages at 98.8\%\ci{97.9}{99.6}, with 7 villages below 99\%. The bounded threshold is more cautious here. It passes 10.9\% of pages at 99.1\%, passes no page in 7 villages, and passes no page on the numeric fields of task~B.

\section{Metric Details}
\label{app:metrics}

The transcriptions write an absent value as 0 and the prompts ask for an empty one, so a value of only zeros counts as empty for every system. An empty answer thus scores 2.0 (A) and 17.5 (B) on the 200-page samples, mostly because the CS list is nearly always absent. Text is normalised to Unicode NFC, Bengali digits are mapped to Western digits, and whitespace is collapsed. Numbers are compared after removing spaces, and a leading-dot decimal such as ``.09'' is read as ``0.09''. A value made only of zeros and dots, such as ``0'', ``00'' or ``.00'', is read as empty in every field. Plot lists are compared as sets of normalised numbers, after splitting any list item on the Arabic thousands separator (U+066C), which some transcriptions use between plot numbers. ANLS follows \citet{biten2019anls} with threshold 0.5. For the edit rate, a plot list is written as its numbers in sorted order joined by commas, so order does not count. Characters are Unicode code points. Each field's edits are capped at its reference length, so a field never costs more than retyping it, with at least one edit for clearing a spurious value. A page with no usable output counts as fully retyped. Edits and reference characters are summed over all fields and pages before dividing. Field confidence is the softmax probability of each generated token from the unmodified model output, before the repetition penalty is applied. The 200-page samples are stratified by village and record type and selected by a deterministic hash. They hold 13--14 pages per village on task~A and 5--25 on task~B. Paired bootstrap intervals use 1,000 resamples of mouzas; the CIs for the gate use 2,000.

\section{Timing Study Details}
\label{app:timing}

Two operators typed the first half and checked the second, and the other two did the reverse. The two ways of working alternated in blocks of five, after six untimed practice pages. A web application recorded each page's active time and submitted values.

\end{document}